# Approximate Planning for Factored POMDPs using Belief State Simplification

**David A. McAllester and Satinder Singh**
AT&T Labs-Research
180 Park Avenue
Florham Park, NJ 07932-0971
{dmac,baveja}@research.att.com

## Abstract

We are interested in the problem of planning for factored POMDPs. Building on the recent results of Kearns, Mansour and Ng, we provide a planning algorithm for factored POMDPs that exploits the accuracy-efficiency tradeoff in the belief state simplification introduced by Boyen and Koller.

## 1 INTRODUCTION

A large number of problems of sequential decision-making in uncertain environments from artificial intelligence, operations-research, and control can be modeled as Markov decision processes (MDPs) [7, 1]. In such problems, at each time step the agent observes the state of the environment and then executes an action causing a reward for the agent and a stochastic transition in the state of the environment. In a finite-horizon MDP, the goal is to choose actions so as to maximize the expected sum of rewards up to the given horizon. A number of algorithms for solving MDPs are available [7, 1]. The complexity of such algorithms is typically some low-order polynomial in the number of states in the environment and of the decision-making horizon [1].

Unfortunately, in many real-world problems the agent's sensors provide only partial information about the state of the environment. This partial information is called an observation and is generated stochastically from the state. Such problems are modeled as *partially observable* MDPs or POMDPs [4, 6]. In a finite-horizon POMDP the goal is again to choose actions so as to maximize the expected sum of rewards up to the given horizon. At each point in time the sequence of observations made by the agent determines a probability distribution over states of the environment. Such a probability distribution is called a belief state. It is well known that the problem of planning in an arbitrary POMDP can be reduced to the problem of planning in the corresponding "belief state" MDP. The number of belief-states is infinite and therefore it is not possible to iterate over belief-states as is normally done in MDP algorithms such as value iteration. However, for a finite-horizon POMDP, various computable forms of value iteration have been defined using piecewise linear representations of value functions [6, 4]. Such algorithms yield a representation of a policy that assigns actions to all belief-states. However, their worst-case complexity is *doubly-exponential* in the horizon.

Note that if the number of observations and actions is finite, then for a finite-horizon problem with a given initial belief state only a finite number of belief-states are reachable. This leads to a straightforward algorithm for selecting actions at reachable belief-states whose complexity is only singly exponential in the horizon. This algorithm is best formalized by viewing a POMDP as a game between the agent and the environment. First the agent selects an action and then the environment randomly selects the next observation. The best action can be selected by searching this game tree to the given horizon. Associated with each node in the game tree is a belief state. Computing the belief state resulting from a given action requires $O(|S|^2)$ operations (see, e.g., [4]), where $S$ is the set of possible states of the environment. The number of nodes in the tree is $O[(|A||O|)^H]$, where $A$ is the set of actions, $O$ is the set of possible observations, and $H$ is the given horizon, i.e., $H$ is the number of allowed actions by the agent. This gives a total run time of $O[|S|^2(|A||O|)^H]$. We can not expect to do better than singly exponential in the horizon because it has been shown that, for horizons polynomial in the size of the POMDP, it is PSPACE hard to determine if the agent can achieve a particular expected total reward [5].

Although we cannot expect to do better than singly-exponential in the horizon, dependence on $|S|$ and $|O|$ can be significantly improved. First, we consider the dependence on $|O|$. In the game between the agent and the environment it is possible to approximately compute the expected reward achievable from a given action by stochastically sampling the observation "responses" a sufficient number of times. This yields an algorithm that achieves an expected total reward no less than $\delta$ from optimal and whose complexity is $O[|S|^2(poly(\frac{1}{\delta})|A|)^H]$. This represents a significant improvement when $|O|$ is large. This approach was developed by Kearns et al. for the case of MDPs and we apply it here to the case of POMDPs.

Next we consider the dependence on $|S|$. To see the significance of the dependence on $|S|$ consider the case of a factored POMDP, i.e., a POMDP in which the state of the environment is described by the values of a set of state variables. The number of states, i.e., $|S|$, is then exponential in the number of state variables. For example, if each state variable has two possible values, and the number of state variables is $n$, then $|S|$ is $2^n$ and so the running time of the above algorithm is $O[2^{2n}(poly(\frac{1}{\delta})|A|)^H]$. For simplicity, from here on we assume that the state variables have only two possible values. While the sampling algorithm can be used with arbitrarily large observation sets, it is restricted to factored POMDPs with a small number of state variables. The dependence on $|S|$ (or $n$) can be improved by associating each node in the search tree with a more efficiently computable simplified belief state approximating the true belief state at that node. One straightforward form of belief state simplification is to approximate each belief state, i.e., a distribution on $2^n$ states, by its product of marginals. Note that simply representing an arbitrary belief state requires $2^n-1$ parameters while a product of marginals state can be represented with $n$ parameters. Unfortunately, the standard product-of-marginals simplification incurs an error that grows with the number of state variables.

Following Boyen and Koller [2] we note that a more realistic simplification processes divides the state variables into a bounded number of classes and then uses a product of marginals simplification which treats each class of variables as a single "metavariable" with an exponential number of values. Under such simplification, representing a single belief state still requires an exponentially large data structure. However, $k$-class belief state simplification cuts the exponent by a factor of $k$. Furthermore, Boyen and Koller show that though the difference between the simplified belief state and the true belief state can grow over time, in rapidly mixing POMDPs the expectation of this difference remains bounded. As our main result, we give a precise analysis relating the quality of belief state approximation to the quality of decision-making.

In summary, in this paper we consider some fixed but arbitrary belief state simplifier and associated notion of simplified belief state. The choice of simplifier, e.g., the choice of the number of variable classes, should reflect a trade-off between the accuracy of simplification and the space and time required to represent and manipulate simplified states. We state the accuracy of our algorithm as a function of the accuracy of simplification. The run time of our algorithm is the time required for a certain number of operations on simplified belief states. Hence our algorithm inherits the accuracy-efficiency trade-off inherent in the choice of the simplifier.

## 2 Problem Definition and Notation

We now define some notation for MDPs. The initial state is denoted by $s_0$. The probability of a transition to state $s'$ conditional upon taking action $a$ in state $s$ is denoted $P(s'|a,s)$. We write $R_s \leq R_{\max}$ for the reward in state $s$.

Following Kearns et al. we work with a discount factor rather than a finite horizon. We take the agent's goal to be that of selecting actions so as to maximize the expected discounted summed reward, i.e., the expected value of the infinite sum $\sum_{t=0}^{\infty} \gamma^t r_t$ where $r_t$ is the reward at time step $t$, and $0 \leq \gamma < 1$ is a discount factor. This sum can be no larger than $\sum_{t=0}^{\infty} \gamma^t R_{\max}$ which equals $\frac{R_{\max}}{1-\gamma}$. The quantity $\frac{1}{1-\gamma}$ is analogous to a horizon time — rewards that occur significantly later than $\frac{1}{1-\gamma}$ have no significant impact on the discounted summed reward and our running time will be exponential in $\frac{1}{1-\gamma}$.

A POMDP consists of an underlying MDP plus a set of possible observations. The probability of observation $o$ when the underlying state is $s$ is denoted $P(o|s)$. For technical convenience we assume the agent knows the initial state and so we can ignore any initial observation.

Here we associate a POMDP with two MDPs — the true belief state MDP and a simplified belief state MDP. The simplified belief state MDP is an approximation of the true belief state MDP using simplified belief states which can be computed and manipulated more efficiently. Both the true belief state MDP and the simplified belief state MDP can be viewed as a game between the agent and an environment. The games associated with the true belief state MDP and the simplified belief state MDP differ in the expected reward and the probability distributions over the next observation at each node in the tree. To precisely compare these two MDPs it is technically convenient to use the game positions as states. A game position where the agent is to select the next action (a move for the agent) is defined by a *history*, i.e., a sequence of action observation pairs.

We now formally define the true belief state MDP $M$ and the the simplified belief state MDP $\hat{M}$. Both $M$ and $\hat{M}$ have histories (game positions) as states. In both cases each history is associated with a belief state, i.e., a probability distribution over states of the underlying MDP. $M$ uses the true belief state $\beta$ defined as follows where $\delta(s_0)$ denotes the probability distribution in which all probability mass is concentrated at the single state $s_0$:

$$\beta(\emptyset) = \delta(s_0), \quad \beta(\sigma; <a,o>) = \mathsf{U}(\beta(\sigma), <a,o>),$$

$$\begin{aligned}\mathsf{U}(\phi, <a,o>)(s') &\equiv P(s'|\phi, <a,o>) \\ &= \frac{1}{P(o|a,\phi)} \sum_{s \in S} \phi(s) P(s'|a,s) P(o|s')\end{aligned}$$

where $P(o|a,\phi) = \sum_{s'} \sum_s \phi(s) P(s'|a,s) P(o|s')$. $\hat{M}$ uses a simplified mapping $\hat{\beta}$ defined as follows:

$$\hat{\beta}(\emptyset) \equiv \mathsf{S}(\delta(s_0)), \quad \hat{\beta}(\sigma; <a,o>) \equiv \mathsf{S}(\mathsf{U}(\hat{\beta}(\sigma), <a,o>)) \quad (1)$$

where $\mathsf{S}$ is a given mapping from belief-states to simplified belief-states — a belief state $\phi$ will be called *simplified* if $\mathsf{S}(\phi) = \phi$.

Now we define transition probabilities $P$ and $\hat{P}$ and reward function $R$ and $\hat{R}$ for $M$ and $\hat{M}$ respectively. For each history $\rho$ and for each action $a$, the history gets extended by $<a,o>$ where $o$ is sampled according to $P(\cdot|a,\rho)$ and $\hat{P}(\cdot|a,\rho)$ as follows:

$$P(o|a,\rho) = \sum_s \beta(\rho)(s) \sum_{s'} P(s'|a,s) P(o|s'),$$

$$\hat{P}(o|a,\rho) = \sum_s \hat{\beta}(\rho)(s) \sum_{s'} P(s'|a,s) P(o|s').$$

The reward functions are defined as follows: $R_\rho = \sum_s \beta(\rho)(s) R_s$, and $\hat{R}_\rho = \sum_s \hat{\beta}(\rho)(s) R_s$.

A policy is a function from agent-move positions in the game tree (histories) to agent actions. Although the optimal policy for an MDP is deterministic, the Kearns et al. algorithm described in the following section is based on stochastic sampling and hence we need to accommodate stochastic policies, i.e., policies which associate each position (history) with a probability distribution over the next action. The expected return starting from history $\rho$ and following stochastic policy $\mu$ is denoted as $V^\mu(\rho)$ in $M$ and $\hat{V}^\mu(\rho)$ in $\hat{M}$. Adapting standard MDP definitions we have that the following recursive relationships are true for all $\rho$.

$$V^\mu(\rho) = R(\beta(\rho)) + \gamma \sum_{<a,o>} P(a|\mu,\rho) P(o|a,\rho) V^\mu(\rho; <a,o>)$$

$$\hat{V}^\mu(\rho) = R(\hat{\beta}(\rho)) + \gamma \sum_{<a,o>} \hat{P}(a|\mu,\rho) \hat{P}(o|a,\rho) \hat{V}^\mu(\rho; <a,o>)$$

Similarly we define optimal value functions $V^*(\rho) = \max_\mu V^\mu(\rho)$, and $\hat{V}^*(\rho) = \max_\mu \hat{V}^\mu(\rho)$.

## 3 The Planning Algorithm

Our algorithm is, with minor adaptations, that of Kearns et al. Our contribution is not the algorithm itself but rather the analysis of the algorithm when applied to the simplified belief state MDP $\hat{M}$. This analysis is given in later sections. Here we describe the algorithm and state the basic results of Kearns et al.

The algorithm defines a stochastic policy, which we denote by $A^\delta$, which takes a history and computes an action. Given $\delta$, $\gamma$ and $R_{\max}$ the algorithm computes a horizon $H$ and a sample size $C$ in a manner described below. Then given a history the algorithm searches the game tree below the node defined by that history to an additional depth of $H$. The algorithm searches all possible agent actions at each agent move and samples $C$ observations at each environment move. At each agent-move position $\rho$ the algorithm computes a value $Q(\rho, a, d)$ for each possible action $a$ using the following relations.

$Q(\rho, a, d) \equiv$

$$\begin{cases} \hat{R}(\rho) & \\ \quad \text{if } d = 0 & \\ \hat{R}(\rho) + \gamma \frac{1}{C} \sum_{o \in \mathcal{O}(\rho,a)} \max_b Q(\rho; <a,o>, b, d-1) & \\ \quad \text{otherwise} & \end{cases}$$

Here, $\mathcal{O}(\rho, a)$ is a set of $C$ samples of observations from $\hat{P}(o|\rho, a)$. The algorithm then selects the root action to be $\text{argmax}_b Q(\rho, b, H)$. Kearns et al. prove the following theorem.

**Theorem 1 (Kearns et al.)** *If*

$$\begin{aligned} H &= \left\lceil \frac{1}{1-\gamma} \ln \frac{4\lambda}{(1-\gamma)^3} \right\rceil \\ C &= \left\lceil \frac{4\lambda^2}{(1-\gamma)^6} (2H \log \frac{4|A|H\lambda^2}{(1-\gamma)^4} + \log \frac{4\lambda}{1-\gamma}) \right\rceil \\ \lambda &= \frac{R_{\max}}{\delta} \end{aligned}$$

*then* $|\hat{V}^{A^\delta}(\rho) - \hat{V}^*(\rho)| \leq \delta$.

Note that the number of belief states computed by the algorithm is the number of nodes in the tree searched which is $O[(|A|C)^H]$.

## 4 Summary of the Analysis

Our main results are two analyses of the policy $A^\delta$ computed by the algorithm of the preceding section. The first analysis, given in section 5, shows that under accurate belief state simplification, and for a rapidly mixing POMDP, a near-optimal policy in $\hat{M}$ performs near-optimally in $M$ indefinitely into the future. The second analysis, given in section 6 shows that independent of the mixing rate of the POMDP, under accurate simplification the policy $\mathcal{A}^\delta$ is near-optimal for $M$ at the beginning.

To state our results we need a measure of the accuracy of simplification and the mixing rate of the POMDP. First, recall that the KL-divergence between two distributions $\Psi$ and $\Phi$, denoted $D(\Phi||\Psi)$ is defined by $D(\Phi||\Psi) = \sum_x \Phi(x) \log_2 \frac{\Phi(x)}{\Psi(x)}$. Boyen and Koller consider belief state simplification satisfying a certain approximation property. Intuitively, we would like to say that the simplification function $S$ has the property that for any belief state $\phi$ we have that $S(\phi)$ is near $\phi$. However, because KL-divergence does not satisfy the triangle inequality we follow Boyen and Koller in assuming that for certain pairs of belief states $\phi$ and $\psi$ we have that $D(\psi||S(\phi)) - D(\psi||\phi)$ is small. More precisely, we take $\psi$ to be $\beta(\rho; <a, o>)$ and $\phi$ to be $U(\hat{\beta}(\rho), <a, o>)$. We say that $S$ is $KL$-$\epsilon$-approximate for $M$ if for all histories $\rho$ and action-observation pairs $<a, o>$ we have the following.

$$D(\beta(\rho; <a, o>)||\hat{\beta}(\rho; <a, o>)) - D(\beta(\rho)||U(\hat{\beta}(\rho), <a, o>)) \leq \epsilon$$

Recall that the $\mathcal{L}_1$-distance between two distributions $\Phi$ and $\Psi$, denoted $||\Phi - \Psi||_1$, is defined by $||\Phi - \Psi||_1 = \sum_x |\Phi(x) - \Psi(x)|$. A simplifier $S$ will be called $\mathcal{L}_1$-$\epsilon$-approximate if $||\delta(s_0) - S(\delta(s_0))||_1 \leq \epsilon$ and for all simplified belief states $\phi$ we have $||U(\phi) - S(U(\phi))||_1 \leq \epsilon$. Following Boyen and Koller, we will call a POMDP $\eta$-mixing if for any two underlying states $s_1$ and $s_2$ and any action $a$ we have that $\sum_{s_3} \min(P(s_3|a, s_1), P(s_3|a, s_2)) \geq \eta$, or equivalently, $||P(s|a, s_1) - P(s|a, s_2)||_1 \leq 2 - 2\eta$.

Now we can state the results of the two analyses of $\mathcal{A}^\delta$.

**Theorem 2** *(Tracking Near Optimality) For $KL$-$\epsilon$-approximate simplification and any $\eta$-mixing POMDP, we have that for all $t \geq 0$,*

$$E^{\mathcal{A}^\delta}_{|\rho|=t}|V^{\mathcal{A}^\delta}(\rho) - V^*(\rho)| \leq \delta + \frac{3R_{\max}}{(1-\gamma)^2}\sqrt{\frac{2\epsilon}{\eta}},$$

*where the expectation is taken over histories generated by running $\mathcal{A}^\delta$ in the true POMDP.*

**Theorem 3** *(Drifting Near Optimality) For $\mathcal{L}_1$-$\epsilon$-approximate simplification and any POMDP we have the following for all $t \geq 0$.*

$$E^{\mathcal{A}^\delta}_{|\rho|=t}|V^{\mathcal{A}^\delta}(\rho) - V^*(\rho)| \leq \delta + \frac{12\epsilon R_{\max}}{(1-\gamma)^3} + \frac{12\epsilon R_{\max} t}{(1-\gamma)^2}$$

*The expectation is taken as in theorem 2.*

## 5   Tracking Near Optimality

In this section we prove theorem 2 which states, in essence, that for accurate belief state simplification, and rapidly mixing POMDPs, the policy $\mathcal{A}^\delta$ is near-optimal. The main component of the analysis is a value transfer lemma stating that for accurate simplification and rapidly mixing MDPs we have that for any policy $\mu$ the simplified value $\hat{V}^\mu(\rho)$ is near the true value $V^\mu(\rho)$ (under expectation over $\rho$). Most of this section involves the proof of this value transfer lemma.

Our departure point is the tracking theorem of Boyen and Koller stated below. In the following theorem, and throughout the remainder of this paper, the expectations are taken over histories $\rho$ generated by running $\mu$ in the true belief state MDP (using $P(o|a, \rho)$ rather than $\hat{P}(o|a, \rho)$).

**Theorem 4** *(Boyen&Koller 98) For any $\eta$-mixing POMDP, any $KL$-$\epsilon$-approximate simplifier for that POMDP, any $t \geq 0$, and any policy $\mu$, we have the following.*

$$E^\mu_{|\rho|=t} D(\beta(\rho)||\hat{\beta}(\rho)) \leq \frac{\epsilon}{\eta}$$

This theorem bounds the expected KL-divergence from the true belief state to the approximate belief state. Our first step is to convert this statement about KL-divergence into a statement about $\mathcal{L}_1$-distance.

**Lemma 5** *For any $\eta$-mixing POMDP, any $KL$-$\epsilon$-approximate $S$ for that POMDP, any $t \geq 0$, and any policy $\mu$ we have the following.*

$$E^\mu_{|\rho|=t}||\beta(\rho) - \hat{\beta}(\rho)||_1 \leq \sqrt{\frac{2\epsilon}{\eta}}$$

**Proof:** For any two distributions $P$ and $Q$ we have the following [3].

$$D(P||Q) \geq \frac{1}{2}(||P - Q||_1)^2$$

This implies the following.

$$\begin{aligned} \frac{\epsilon}{\eta} &\geq E^\mu_{|\rho|=t} D(\beta(\rho)||\hat{\beta}(\rho)) \\ &\geq \frac{1}{2} E^\mu_{|\rho|=t}(||\beta(\rho) - \hat{\beta}(\rho)||_1)^2 \\ &\geq \frac{1}{2}(E^\mu_{|\rho|=t}||\beta(\rho) - \hat{\beta}(\rho)||_1)^2 \end{aligned}$$

Which implies the lemma.    □

Our objective is to bound the true value of $\mathcal{A}^\delta$, i.e., the value under $P(o|a, \rho)$ and $R_\rho$ rather than $\hat{P}(o|a, \rho)$ and $\hat{R}_\rho$. The next step is to bound the (expected) difference between these fundamental quantities.

**Lemma 6** *For any $\eta$-mixing POMDP, any $KL$-$\epsilon$-approximate $S$ for that POMDP, any policy $\mu$, and*

any $t \geq 0$:

$$E^\mu_{|\rho|=t} ||P(o|a,\rho) - \hat{P}(o|a,\rho)||_1 \leq \sqrt{\frac{2\epsilon}{\eta}}.$$

**Proof:** Note that For any $\rho$

$$\begin{aligned}
&||P(o|a,\rho) - \hat{P}(o|a,\rho)||_1 \\
&= \sum_o |\sum_s (\beta(\rho)(s) - \hat{\beta}(\rho)(s)) \sum_{s'} P(s'|a,s) P(o|s')| \\
&\leq \sum_s \sum_o |(\beta(\rho)(s) - \hat{\beta}(\rho)(s))| \sum_{s'} P(s'|a,s) P(o|s') \\
&= \sum_s |\beta(\rho)(s) - \hat{\beta}(\rho)(s)| \sum_o \sum_{s'} P(s'|a,s) P(o|s') \\
&= \sum_s |\beta(\rho)(s) - \hat{\beta}(\rho)(s)| \\
&= ||\beta(\rho) - \hat{\beta}(\rho)||_1
\end{aligned}$$

which together with lemma 5 implies the result. □

**Lemma 7** *For any $\eta$-mixing POMDP, any KL-$\epsilon$-approximate S for that POMDP, any policy $\mu$, and any $t \geq 0$:*

$$E^\mu_{|\rho|=t} |R_\rho - \hat{R}_\rho| \leq R_{\max} \sqrt{\frac{2\epsilon}{\eta}}.$$

**Proof:** For any $\rho$

$$\begin{aligned}
|R_\rho - \hat{R}_\rho| &= |\sum_s (\beta(\rho)(s) - \hat{\beta}(\rho)(s)) R_s| \\
&\leq \sum_s |(\beta(\rho)(s) - \hat{\beta}(\rho)(s))| R_{\max} \\
&= R_{\max} ||\beta(\rho) - \hat{\beta}(\rho)||_1
\end{aligned}$$

which together with lemma 5 implies the result. □

We can now prove the value transfer lemma.

**Lemma 8** *(Value Transfer Lemma) For any $\eta$-mixing POMDP, any KL-$\epsilon$-approximate S for that POMDP, and any policy $\mu$ we have the following.*

$$E^\mu_{|\rho|=t} |V^\mu(\rho) - \hat{V}^\mu(\rho)| \leq \frac{R_{\max}}{(1-\gamma)^2} \sqrt{\frac{2\epsilon}{\eta}}.$$

**Proof:** Define $\Delta_t$ to be $E^\mu_{|\rho|=t} |V^\mu(\rho) - \hat{V}^\mu(\rho)|$, and $\Delta$ to be $\max_t \Delta_t$. For all $t$ we have the following.

$$\begin{aligned}
\Delta_t &= E^\mu_{|\rho|=t} |V^\mu(\rho) - \hat{V}^\mu(\rho)| \\
&\leq E^\mu_{|\rho|=t} |R(\rho) - \hat{R}(\rho)| \\
&\quad + \gamma E^\mu_{|\rho|=t} \sum_a P(a|\mu,\rho) \left| \begin{array}{l} \sum_o P(o|a,\rho) V^\mu(\rho;<a,o>) \\ -\sum_o \hat{P}(o|a,\rho) \hat{V}^\mu(\rho;<a,o>) \end{array} \right| \\
&\leq R_{\max} \sqrt{\frac{2\epsilon}{\eta}} \\
&\quad + \gamma E^\mu_{|\rho|=t} \sum_a P(a|\mu,\rho) \sum_o P(\bullet|a,\rho) \left| \begin{array}{l} V^\mu(\rho;<a,o>) \\ -\hat{V}^\mu(\rho;<a,o>) \end{array} \right| \\
&\quad + \gamma E^\mu_{|\rho|=t} \sum_a P(a|\mu,\rho) \sum_o \hat{V}^\mu(\rho;<a,o>) \left| \begin{array}{l} P(o|a,\rho) \\ -\hat{P}(o|a,\rho) \end{array} \right|
\end{aligned}$$

$$\begin{aligned}
&\leq R_{\max} \sqrt{\frac{2\epsilon}{\eta}} + \gamma \Delta \\
&\quad + \gamma V_{\max} E^\mu_{|\rho|=t} \sum_a P(a|\mu,\rho) \sum_o \left| \begin{array}{l} P(o|a,\rho) \\ -\hat{P}(o|a,\rho) \end{array} \right| \\
&\leq R_{\max} \sqrt{\frac{2\epsilon}{\eta}} + \gamma \Delta + \gamma V_{\max} \sqrt{\frac{2\epsilon}{\eta}} \\
&= V_{\max} \sqrt{\frac{2\epsilon}{\eta}} + \gamma \Delta
\end{aligned}$$

Therefore $\Delta \leq V_{\max} \sqrt{\frac{2\epsilon}{\eta}} + \gamma \Delta$, which implies that $\Delta \leq \frac{V_{\max}}{1-\gamma} \sqrt{\frac{2\epsilon}{\eta}} = \frac{R_{\max}}{(1-\gamma)^2} \sqrt{\frac{2\epsilon}{\eta}}$. □

**Corollary 9** *For any $\eta$-mixing POMDP, any KL-$\epsilon$-approximate S for that POMDP, and any policies $\mu_1$ and $\mu_2$:*

$$E^{\mu_1}_{|\rho|=t} |V^{\mu_2}(\rho) - \hat{V}^{\mu_2}(\rho)| \leq \frac{R_{\max}}{(1-\gamma)^2} \sqrt{\frac{2\epsilon}{\eta}}.$$

**Proof:** Let $\mu_3$ be the policy defined by $\mu_3(\rho) = \mu_1(\rho)$ if $|\rho| < t$ and $\mu_2(\rho)$ otherwise. We now have the following.

$$E^{\mu_1}_{|\rho|=t} |V^{\mu_2}(\rho) - \hat{V}^{\mu_2}(\rho)| = E^{\mu_3}_{|\rho|=t} |V^{\mu_3}(\rho) - \hat{V}^{\mu_3}(\rho)|$$

The result now follows from lemma 8 applied to $\mu_3$. □

**Lemma 10** *For any $\eta$-mixing POMDP, any KL-$\epsilon$-approximate S for that POMDP, any $t \geq 0$, and any policy $\mathcal{A}$ we have the following.*

$$E^\mathcal{A}_{|\rho|=t} |\hat{V}^*(\rho) - V^*(\rho)| \leq \frac{2R_{\max}}{(1-\gamma)^2} \sqrt{\frac{2\epsilon}{\eta}}$$

**Proof:** Let $\mu^*$ be the optimal policy for the belief state MDP and let $\hat{\mu}^*$ be the optimal policy for the simplified belief state MDP. If $V^*(\rho) \geq \hat{V}^*(\rho)$ then $V^{\mu^*}(\rho) \geq \hat{V}^{\hat{\mu}^*}(\rho) \geq \hat{V}^{\mu^*}(\rho)$ and so $|V^*(\rho) - \hat{V}^*(\rho)| \leq |V^{\mu^*}(\rho) - \hat{V}^{\mu^*}(\rho)|$. Similarly, for $V^*(\rho) < \hat{V}^*(\rho)$ we have $|V^*(\rho) - \hat{V}^*(\rho)| \leq |V^{\hat{\mu}^*}(\rho) - \hat{V}^{\hat{\mu}^*}(\rho)|$. In either case we have the following.

$$|V^*(\rho) - \hat{V}^*(\rho)| \leq |V^{\hat{\mu}^*}(\rho) - \hat{V}^{\hat{\mu}^*}(\rho)| + |V^{\mu^*}(\rho) - \hat{V}^{\mu^*}(\rho)|$$

We now get the desired result by taking the expectation over $\rho$ and bounding the resulting expectation of the right hand side above using corollary 9. □

Finally, we can prove theorem 2 by noting that $E^{\mathcal{A}^\delta}_{|\rho|=t} |V^{\mathcal{A}^\delta}(\rho) - V^*(\rho)|$ can be no greater than the sum of $E^{\mathcal{A}^\delta}_{|\rho|=t} |V^{\mathcal{A}^\delta}(\rho) - \hat{V}^{\mathcal{A}^\delta}(\rho)|$, $E^{\mathcal{A}^\delta}_{|\rho|=t} |\hat{V}^{\mathcal{A}^\delta}(\rho) - \hat{V}^*(\rho)|$, and $E^{\mathcal{A}^\delta}_{|\rho|=t} |\hat{V}^*(\rho) - V^*(\rho)|$.

## 6  Drifting Near Optimality

In this section we prove theorem 3 which states, in essence, that for accurate belief state simplification,

and for any POMDP, the policy $\mathcal{A}^\delta$ is near-optimal near the beginning. Without the assumption of rapid mixing it is possible that errors due to simplification accumulate with time. However, in the discounted case studied here the value of a given history is only sensitive to errors within an effective planning horizon determined by the discount factor. We first bound how rapidly errors due to simplification can accumulate and then use this bound to prove the value transfer lemma.

In the appendix we prove the following lemma which is analogous to that of Boyen and Koller except that it gives a bound on $\mathcal{L}_1$-distance rather than KL-divergence and the bound increases with time.

**Lemma 11** *Under $\mathcal{L}_1$-$\epsilon$-approximate simplification we have that for any (possibly unmixing) POMDP, for all $t$, and for all $\mu$,*

$$E^\mu_{|\rho|=t}||\beta(\rho) - \hat{\beta}(\rho)||_1 \leq 4\epsilon(t+1)$$

As in Section 5 the above bound on $\mathcal{L}_1$-distance yields bound on $\hat{R}$ and $\hat{P}$, namely $E^\mu_{|\rho|=t}|R_\rho - \hat{R}_\rho| \leq 4\epsilon t R_{max}$, and $E^\mu_{|\rho|=t}||P^\mu_\rho - \hat{P}^\mu_\rho||_1 \leq 4\epsilon t$.

Next we prove the value transfer lemma.

**Lemma 12**
*(Value Transfer Lemma) For any POMDP, and for any $\epsilon$-approximate S for that POMDP, for any policy $\mu$:*

$$E^{\mathcal{A}^\delta}_{|\rho|=t}|V^\mu(\rho) - \hat{V}^\mu(\rho)| \leq \frac{4\epsilon V_{max}}{(1-\gamma)^2} + \frac{4\epsilon V_{max} t}{1-\gamma}$$

**Proof:** Let $\Delta_t \equiv E^\mu_{|\rho|=t}|V^\mu(\rho) - \hat{V}^\mu(\rho)|$. First we will show that

$$\Delta_t \leq 4\epsilon(t+1)V_{max} + \gamma\Delta_{t+1} \quad (2)$$

$$\begin{aligned}
\Delta_t &= E^\mu_{|\rho|=t}|V^\mu(\rho) - \hat{V}^\mu(\rho)| \\
&\leq E^\mu_{|\rho|=t}|R(\beta(\rho)) - R(\hat{\beta}(\rho))| \\
&\quad + E^\mu_{|\rho|=t}\gamma \left| \begin{array}{l} \sum_{<a,o>} P_\rho(<a,o>)V^\mu(\rho;<a,o>) \\ -\sum_{<a,o>} \hat{P}_\rho(<a,o>)\hat{V}^\mu(\rho;<a,o>) \end{array} \right| \\
&\leq 4\epsilon(t+1)R_{max} \\
&\quad + \gamma E^\mu_{|\rho|=t} \sum_{<a,o>} P_\rho(<a,o>) \left| \begin{array}{l} V^\mu(\rho;<a,o>) \\ -\hat{V}^\mu(\rho;<a,o>) \end{array} \right| \\
&\quad + E^\mu_{|\rho|=t}\gamma \sum_{<a,o>} \hat{V}^\mu(\rho;<a,o>) \left| \begin{array}{l} P_\rho(<a,o>) \\ -\hat{P}_\rho(<a,o>) \end{array} \right| \\
&\leq 4\epsilon(t+1)R_{max} + \gamma\Delta_{t+1} \\
&\quad + \gamma V_{max} E^\mu_{|\rho|=t} \sum_{<a,o>} \left| \begin{array}{l} P_\rho(<a,o>) \\ -\hat{P}_\rho(<a,o>) \end{array} \right|
\end{aligned}$$

$$\begin{aligned}
&\leq 4\epsilon(t+1)R_{max} + \gamma\Delta_{t+1} + \gamma 4\epsilon(t+1)V_{max} \\
&= 4\epsilon(t+1)V_{max} + \gamma\Delta_{t+1}
\end{aligned}$$

Now we show that $\Delta_t \leq 4\epsilon(t+1)V_{max} + \gamma\Delta_{t+1}$ implies that $\Delta_t \leq \frac{4\epsilon V_{max} t}{1-\gamma} + \frac{4\epsilon V_{max}}{(1-\gamma)^2}$.

Plugging $\Delta_{t+1} \leq V_{max}$ into Equation 2 we get $\Delta_t \leq 4\epsilon(t+1)V_{max} + \gamma V_{max}$. By continuing this unfolding in the limit we get that $\Delta_t \leq 4\epsilon V_{max} \sum_{i=0}^\infty \gamma^i(t+1+i)$. Summing this series gives $\Delta_t \leq \frac{4\epsilon V_{max} t}{1-\gamma} + \frac{4\epsilon V_{max}}{(1-\gamma)^2}$. □

The proofs of corollary 9 and lemma 10 can be used here to show that for all $t \geq 0$ we have the following.

$$E^{\mathcal{A}^\delta}_{|\rho|=t}|V^*(\rho) - \hat{V}^*(\rho)| \leq \frac{8\epsilon V_{max}}{(1-\gamma)^2} + \frac{8\epsilon V_{max} t}{1-\gamma}$$

As with theorem 2, theorem 3 now follows from the observation that $E^{\mathcal{A}^\delta}_{|\rho|=t}|V^{\mathcal{A}^\delta}(\rho) - V^*(\rho)|$ can be no greater than the sum of $E^{\mathcal{A}^\delta}_{|\rho|=t}|V^{\mathcal{A}^\delta}(\rho) - \hat{V}^{\mathcal{A}^\delta}(\rho)|$, $E^{\mathcal{A}^\delta}_{|\rho|=t}|\hat{V}^{\mathcal{A}^\delta}(\rho) - \hat{V}^*(\rho)|$, and $E^{\mathcal{A}^\delta}_{|\rho|=t}|\hat{V}^*(\rho) - V^*(\rho)|$.

## 7  Conclusion and Future Work

We showed that the straightforward application of Kearns et al.'s planning algorithm to POMDPs leads to a more efficient algorithm than existing algorithms for solving POMDPs, at least for problems with large observation sets and tractable belief-state computation. For problems in which exact belief-state computation is too expensive, by building on the Kearns et al. algorithm, and the work on belief state simplification by Boyen and Koller, we have established that the accuracy-efficiency trade-off in belief state simplification can be used to achieve an accuracy efficiency trade-off in planning. Although we do not yet have a reasonable planning algorithm for factored POMDPs that is polynomial in the number of state variables, belief state simplification can in principle dramatically reduce the exponent in the running time of the exponential planning algorithm.

A significant open problem is to find some additional conditions under which truly polynomial (in the number of state variables) POMDP planning can be achieved. It seems possible that if one imposes a "smoothness" criterion on the reward function, e.g., that the reward is a sum of local rewards, then polynomial factored POMDP planning can be done.

## A   Proof of Lemma 11

First we prove some general lemmas about $\mathcal{L}_1$ distance and then prove Lemma 11.

**Lemma 13** *Let $\beta$ and $\hat{\beta}$ be two distributions on the same set $S$ and let $P(x|s)$ be a conditional probability function on $X \times S$, i.e., $P(x|s) \in [0, 1]$ and $\sum_x P(x|s) = 1$ for any fixed $s$. We have the following where $P(x|s)\beta(s)$ denotes the obvious distribution on $X \times S$.*

$$\|P(x|s)\beta(s) - P(x|s)\hat{\beta}(s)\|_1 = \|\beta - \hat{\beta}\|_1$$

**Proof:**

$$\|P(x|s)\beta(s) - P(x|s)\hat{\beta}(s)\|_1$$
$$= \sum_x \sum_s |P(x|s)\beta(s) - P(x|s)\hat{\beta}(s)|$$
$$= \sum_s |\beta(s) - \hat{\beta}(s)| \sum_x P(x|s)$$
$$= \sum_s |\beta(s) - \hat{\beta}(s)|$$
$$= \|\beta - \hat{\beta}\|_1$$

**Lemma 14** *Let $P$ and $Q$ be any two distributions on $X \times Y$. Let $P(x)$ denote the marginal distribution on $X$, i.e., $P(x) = \sum_y P(x, y)$, and similarly for $Q(x)$. We then have the following.*

$$\|P(x) - Q(x)\|_1 \le \|P(x,y) - Q(x,y)\|_1$$

**Proof:**

$$\|P(x) - Q(x)\|_1 = \sum_x |\sum_y P(x,y) - \sum_y Q(x,y)|$$
$$\le \sum_x \sum_y |P(x, y) - Q(x, y)|$$
$$= \|P(x, y) - Q(x, y)\|_1$$

**Lemma 15** *Let $P$ and $Q$ be any two distributions on $X \times O$. Let $P(o)$ denote the marginal distribution on $O$, i.e., $P(o) = \sum_x P(x, o)$, and similarly for $Q(o)$. We then have the following.*

$$E_{o \sim P(o)} \|P(x|o) - Q(x|o)\|_1 \le$$
$$\|P(x,o) - Q(x,o)\|_1 + \|P(o) - Q(o)\|_1$$

**Proof:**

$$E_{o \sim P(o)} \|P(x|o) - Q(x|o)\|_1$$
$$= \sum_o P(o) \sum_x |\frac{P(x, o)}{P(o)} - \frac{Q(x, o)}{Q(o)}|$$
$$\le \sum_o P(o) \sum_x |\frac{P(x, o)}{P(o)} - \frac{Q(x, o)}{P(o)}|$$
$$+ \sum_o P(o) \sum_x |\frac{Q(x, o)}{P(o)} - \frac{Q(x, o)}{Q(o)}|$$
$$= \sum_o \sum_x |P(x, o) - Q(x, o)|$$
$$+ \sum_o P(o) \sum_x |\frac{Q(x, o)}{P(o)} - \frac{Q(x, o)}{Q(o)}|$$
$$= \|P(x, o) - Q(x, o)\|_1$$
$$+ \sum_o P(o) \sum_x Q(x, o)|\frac{1}{P(o)} - \frac{1}{Q(o)}|$$
$$= \|P(x, o) - Q(x, o)\|_1$$
$$+ \sum_o P(o)|\frac{1}{P(o)} - \frac{1}{Q(o)}| \sum_x Q(x, o)$$
$$= \|P(x, o) - Q(x, o)\|_1$$
$$+ \sum_o P(o)|\frac{1}{P(o)} - \frac{1}{Q(o)}|Q(o)$$
$$= \|P(x, o) - Q(x, o)\|_1 + \sum_o |Q(o) - P(o)|$$
$$= \|P(x, o) - Q(x, o)\|_1 + \|Q(o) - P(o)\|_1$$

We need one more lemma before proving Lemma 11. We assume a fixed (stochastic) policy $\mu$ defined by the probabilities $P(a|\rho, \mu)$. We define $P(w, \sigma|\rho, s, t, \mu)$ to be be the probability that if we assume that the hidden state at time $|\rho|$ is $s$ and then run forward for $t$ additional steps in the underlying MDP we generate additional history $\sigma$ and end in final state $w$. If $\beta$ is a belief state we let $P(w, \sigma|\rho, \beta, t, \mu)$ be $\sum_s \beta(s) P(w, \sigma|\rho, s, t, \mu)$. We also let $P(w|\sigma, \rho, \beta, \mu)$ be the probability of $w$ given $\sigma$ under to the joint distribution $P(w, \sigma|\rho, \beta, t, \mu)$.

We say that $\beta_i$ is simplified if $S(\beta_i) = \beta_i$. We say that a belief state $\delta_i$ is pre-simplified if it can be written as $U(<a, o>, \beta_i)$ for some $<a, o>$ and simplified $\beta_i$. We say that $S$ is $\mathcal{L}_1$-$\epsilon$-approximate if for any pre-simplified $\delta_i$ we have $\|S(\delta_i) - \delta_i\|_1 \le \epsilon$.

**Lemma 16** *For any $\mathcal{L}_1$-$\epsilon$-approximate function $S$, pre-simplified belief state $\delta$, history $\rho$, and policy $\mu$ we have the following where the expectation is taken over the marginal on $\sigma$ of the joint distribution defined by $P(w, \sigma|\rho, \delta, t, \mu)$.*

$$E_\sigma \|P(w|\sigma, \rho, \delta, t, \mu) - P(w|\sigma, \rho, S(\delta), t, \mu)\|_1 \le 2\epsilon$$

**Proof:** By lemma 15 we have the following.

$$E_\sigma ||P(w|\sigma,\rho,\delta,t,\mu) - P(w|\sigma,\rho,S(\delta),t,\mu)||_1$$
$$\leq ||P(w,\sigma|\rho,\delta,t,\mu) - P(w,\sigma|\rho,S(\delta),t,\mu)||_1$$
$$+ ||P(\sigma|\rho,\delta,t,\mu) - P(\sigma|\rho,S(\delta),t,\mu)||_1$$

Combining lemmas 13 and 14 we have the following.

$$||P(w,\sigma|\rho,\delta,t,\mu) - P(w,\sigma|\rho,S(\delta),t,\mu)||_1$$
$$\leq ||\delta - S(\delta)||_1 \leq \epsilon$$

$$||P(\sigma|\rho,\delta,t,\mu) - P(\sigma|\rho,S(\delta),t,\mu)||_1 \leq ||\delta - S(\delta)||_1 \leq \epsilon$$

These together give the desired result. □

Let the functions $\beta$ and $\hat{\beta}$ be defined as in the paper and let the expectations be taken over running $\mu$ in the true belief-state MDP.

We now prove Lemma 11 (restated here).

**Lemma 11** If $S$ is $\mathcal{L}_1$-$\epsilon$-approximate then for any $t \geq 0$ and for all $\mu$ we have the following.

$$E^\mu_{|\rho|=t}||\beta(\rho) - \hat{\beta}(\rho)||_1 \leq 4\epsilon(t+1)$$

**Proof:** Note that $P(w|\sigma,\rho,\beta,\mu)$ satisfies the following conditions.

$$P(w|\emptyset,\rho,\beta,\mu) = \beta(w)$$

$$P(w|<a,o>;\sigma,\rho,\beta,\mu)$$
$$= P(w|\sigma,\rho;<a,o>,U(<a,o>,\beta),\mu)$$

This implies that $P(w|\sigma,\rho,\beta,\mu)$ is independent of $\mu$ and $\rho$ and hence can be written as $P(w|\sigma,\beta)$. Let $\hat{P}(w|\sigma,\beta)$ be the estimate of $P(w|\sigma,\beta)$ gotten by running with simplified intermediate belief states. More formally this estimate is defined by the following equations.

$$\hat{P}(w|\emptyset,\beta) = \beta(w)$$

$$\hat{P}(w|<a,o>;\sigma,\beta) = \hat{P}(w|\sigma, S(U(<a,o>,\beta)))$$

We now prove the following general statement for any $\rho$ and simplified belief state $\beta$ and where $E^{\rho,\beta}_{|\sigma|=t}f(\sigma)$ denotes the expectation of $f(\sigma)$ over the histories $\sigma$ defined by $P(\sigma|\rho,\beta,t,\mu)$.

$$\Delta_{\rho,\beta,t} \equiv E^{\rho,\beta}_{|\sigma|=t}||P(w|\sigma,\beta) - \hat{P}(w|\sigma,\beta)||_1 \leq 4\epsilon t \quad (3)$$

The proof is by induction on $t$. For $t = 0$ the result is immediate. Now assume the result for $t$ and consider $t+1$. In the following $\delta$ abbreviates $U(<a,o>,\beta)$ and $P(<a,o>)$ abbreviates $P(<a,o>|\rho,\beta,\mu)$.

$$\Delta_{\rho,\beta,t} = E^{\rho,\beta}_{|\sigma|=t+1}||P(w|\sigma,\beta) - \hat{P}(w|\sigma,\beta)||_1$$
$$= \sum_{<a,o>} P(<a,o>)E^{\rho;<a,o>,\delta}_{|\sigma'|=t}||P(w|\sigma',\delta)$$
$$\quad - \hat{P}(w|\sigma',S(\delta))||_1$$
$$\leq \sum_{<a,o>} P(<a,o>)E^{\rho;<a,o>,\delta}_{|\sigma'|=t}||P(w|\sigma',\delta)$$
$$\quad - P(w|\sigma',S(\delta))||_1$$
$$\quad + \sum_{<a,o>} P(<a,o>)E^{\rho;<a,o>,\delta}_{|\sigma'|=t}||P(w|\sigma',S(\delta))$$
$$\quad - \hat{P}(w|\sigma',S(\delta))||_1$$
$$\leq 2\epsilon + \sum_{<a,o>} P(<a,o>)E^{\rho;<a,o>,\delta}_{|\sigma'|=t}||P(w|\sigma',S(\delta))$$
$$\quad - \hat{P}(w|\sigma',S(\delta))||_1$$
$$= 2\epsilon + \sum_{<a,o>} P(<a,o>)$$
$$\quad \sum_{|\sigma'|=t} P(\sigma|\rho;<a,o>,\delta,\mu)||P(w|\sigma',S(\delta))$$
$$\quad - \hat{P}(w|\sigma',S(\delta))||_1$$
$$\leq 2\epsilon + \sum_{<a,o>} P(<a,o>)$$
$$\quad \sum_{|\sigma'|=t} 2|P(\sigma|\rho;<a,o>,\delta,\mu)$$
$$\quad - P(\sigma|\rho;<a,o>,S(\delta),\mu)|$$
$$\quad + \sum_{<a,o>} P(<a,o>)\sum_{|\sigma'|=t}P(\sigma|\rho;<a,o>,S(\delta),\mu)$$
$$\quad ||P(w|\sigma',S(\delta)) - \hat{P}(w|\sigma',S(\delta))||_1$$
$$= 4\epsilon + \sum_{<a,o>} P(<a,o>)$$
$$\quad E^{\rho;<a,o>,S(\delta)}_{|\sigma'|=t}||P(w|\sigma',S(\delta))$$
$$\quad - \hat{P}(w|\sigma',S(\delta))||_1$$
$$= 4\epsilon + 4\epsilon t = 4\epsilon(t+1)$$

We now show that 3 implies the main theorem. Let $\Pi$ denote the initial belief state that has all its mass on state $s_0$.

$$E^\mu_{|\rho|=t}||\beta(\rho) - \hat{\beta}(\rho)||_1$$
$$= E^{\emptyset,\Pi}_{|\rho|=t}||P(w|\rho,\Pi) - \hat{P}(w|\rho,S(\Pi))||_1$$
$$= \sum_{|\rho|=t} P(\rho|\emptyset,\Pi,t,\mu)||P(w|\rho,\Pi) - \hat{P}(w|\rho,S(\Pi))||_1$$
$$\leq \sum_{|\rho|=t} 2|P(\rho|\emptyset,\Pi,t,\mu) - P(\rho|\emptyset,S(\Pi),t,\mu)|$$
$$+ \sum_{|\rho|=t} P(\rho|\emptyset,S(\Pi),t,\mu)||P(w|\rho,\Pi) - \hat{P}(w|\rho,S(\Pi))||_1$$
$$\leq 2\epsilon + E^{\emptyset,S(\Pi)}_{|\rho|=t}||P(w|\rho,\Pi) - \hat{P}(w|\rho,S(\Pi))||_1$$
$$\leq 2\epsilon + E^{\emptyset,S(\Pi)}_{|\rho|=t}||P(w|\rho,\Pi) - P(w|\rho,S(\Pi))||_1$$
$$\quad + E^{\emptyset,S(\Pi)}_{|\rho|=t}||P(w|\rho,S(\Pi)) - \hat{P}(w|\rho,S(\Pi))||_1$$
$$\leq 3\epsilon + 4\epsilon t \leq 4\epsilon(t+1)$$